\pdfoutput=1

\documentclass[11pt]{article}

\usepackage[preprint]{coling}

\usepackage{times}
\usepackage{latexsym}

\usepackage[T1]{fontenc}

\usepackage[utf8]{inputenc}

\usepackage{microtype}

\usepackage{inconsolata}
\usepackage{amsmath}
\usepackage{makecell}
\usepackage{multirow}
\usepackage{graphicx} 
\usepackage{float} 
\usepackage{subfigure} 
\usepackage{array}
\usepackage{makecell}

\usepackage{xurl}

\title{Enhancing Knowledge Distillation of Large Language Models through Efficient Multi-Modal Distribution Alignment}

\author{
    Tianyu Peng\textsuperscript{1,2,3} \and 
    Jiajun Zhang\textsuperscript{1,2,3,4}\thanks{Corresponding Author.}   \\
    \textsuperscript{1}School of Artificial Intelligence, University of Chinese Academy of Sciences\\
    \textsuperscript{2}Institute of Automation, Chinese Academy of Sciences\\
    \textsuperscript{3}Wuhan AI Research, 
    \textsuperscript{4}Shanghai Artificial Intelligence Laboratory, Shanghai, China\\
    \texttt{\href{pengtianyu2022@ia.ac.cn}{pengtianyu2022@ia.ac.cn}, \href{jjzhang@nlpr.ia.ac.cn}{jjzhang@nlpr.ia.ac.cn}}
}
\begin{document}
\maketitle
\begin{abstract}

Knowledge distillation (KD) is an effective model compression method that can transfer the internal capabilities of large language models (LLMs) to smaller ones. However, the multi-modal probability distribution predicted by teacher LLMs causes difficulties for student models to learn. In this paper, we first demonstrate the importance of multi-modal distribution alignment with experiments and then highlight the inefficiency of existing KD approaches in learning multi-modal distributions. To address this problem, we propose Ranking Loss based Knowledge Distillation (RLKD), which encourages the consistency of the ranking of peak predictions between the teacher and student models. By incorporating word-level ranking loss, we ensure excellent compatibility with existing distillation objectives while fully leveraging the fine-grained information between different categories in peaks of two predicted distribution. Experimental results demonstrate that our method enables the student model to better learn the multi-modal distributions of the teacher model, leading to a significant performance improvement in various downstream tasks.{\footnote{Our code is available at \href{https://github.com/Pty72/RLKD}{https://github.com/Pty72/RLKD}.}}

\end{abstract}

\section{Introduction}

In recent years, large language models (LLMs, \citealt{brown2020language, zeng2023glm130b, openai2023gpt4, touvron2023llama, yang2023baichuan, jiang2024mixtral}) have demonstrated their great power to solve natural language processing tasks. Existing research \cite{kaplan2020scaling, wei2022emergent} shows that language models tend to perform better as the number of parameters increases. However, the training and deployment of large scale models involve high costs, coupled with less usability and flexibility. Therefore, model compression techniques \cite{DBLP:journals/corr/abs-2308-07633} for LLMs are drawing more and more attention.

Knowledge distillation (KD, \citealt{hinton2015distilling}) is one of the representative approaches for model compression. It facilitates efficient knowledge transfer to smaller student models by using the full probability distribution output from teacher models as a guiding signal during optimization. Therefore, designing better distillation objectives that make it more efficient for student models to learn the overall probability distribution of teacher models is the focus in KD research.

Particularly for LLMs, the lengthy and complex probability distribution causes more learning difficulties. Due to the diversity of natural language, the predicted probability distribution of LLMs is often multi-modal (quantitative analysis results are shown in Appendix \hyperref[A]{A}), which contains multiple potential correct predictions for a given input. Consequently, improving the learning of multi-modal distribution becomes the focus of LLMs' KD.

\begin{figure} 
\centering 
\includegraphics[width=0.46\textwidth]{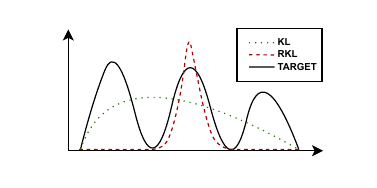} 
\vspace{-0.4cm}
\caption{An theoretical example illustrates the situations that can arise when using KL or RKL as distillation objective to fit multi-modal distribution.} 
\vspace{-0.5cm}
\label{Fig.kl} 
\end{figure}

To achieve this goal, previous studies have already explored several distillation objectives for LLMs. Conventional KD \cite{hinton2015distilling} uses forward Kullback-Leibler divergence (KL) as the optimization objective. However, KL predisposes to the mode-averaging problem \cite{wen2023fdivergence}, whereby student models tend to learn too smooth distributions (as shown in Figure \hyperref[Fig.kl]{1}). Therefore, more recent work \cite{gu2023knowledge, agarwal2024onpolicy} has employed reverse KL (RKL) instead of KL, claiming that this optimization objective can better focus on peak predictions. Nevertheless, optimizing RKL tends to get an overconcentration of the probability predictions of the student model in some specific intervals (as illustrated in Figure \hyperref[Fig.kl]{1}). Then, \newcite{wen2023fdivergence} propose using symmetric divergences as the distillation objective to alleviate mode problems caused by KL and RKL. In contrary, \newcite{DBLP:journals/corr/abs-2404-02657} verify through theory and experiments on toy data that KL and RKL do not suffer from the above issues, and they instead share the same optimization objective in KD of LLMs.

However, despite the demonstrated effectiveness of the distillation objectives proposed in the aforementioned work in fitting multi-modal distributions, either through theoretical or toy experiments, they do not specifically showcase the learning capability of student models for multi-modal distributions. We still have no idea whether these distillation objectives truly enhance the learning ability of student models for multi-modal distributions in real-world tasks, hence making it difficult to ascertain the source of the improvements in downstream tasks.

In order to address the above issues, we propose in this paper Ranking Loss based Knowledge Distillation (RLKD) for LLMs. We first verify the relationship between multi-modal predictions and model performance, and experimentally demonstrate the problems of existing distillation objectives in fitting multi-modal distributions. In response to the identified problems, we introduce the word-level ranking loss, which is based on Spearman's rank correlation coefficient (SRCC), to optimize the degree of consistency in the order of peak predictions between the teacher and student models. In particular, we convert the learning of multi-modal distributions during the KD process into the learning of the top-k sampling \cite{holtzman2020curious} order. Through ranking loss, we ensure excellent compatibility with existing distillation objectives while fully leveraging the fine-grained information between different categories predicted by two distribution peaks. Additionally, we verify and demonstrate through real-scenario experiments the impact of introducing ranking loss into KD on the learning ability of student models for multi-modal distributions.

Experimental results indicate that the quality of multi-modal predictions is closely related to the performance of the model, while existing distillation objectives lack the ability to fit multi-modal distributions effectively. Subsequently, our proposed method effectively enhances the student model's learning ability to predict multi-modal distributions during the distillation process and exhibits good compatibility with existing distillation objectives. We also validate the ranking loss on diverse datasets from multiple tasks, showing significant improvements in downstream KD tasks.

In general, our main contributions are as follows:
\begin{itemize}
    \item [1.]
    We propose a word-level ranking loss for KD of LLMs, that significantly improves the student model's multi-modal distribution learning ability and performance on downstream tasks.
    \item [2.]
    We analyze the importance of multi-modal distribution alignment through experiments. Additionally, we verify the shortcomings of existing methods in peak prediction learning and achieve significant improvements with our proposed method.
\end{itemize}

\section{Related Work}

\subsection{KD of LLMs}

Nowadays, many LLMs are no longer open source due to commercial and other considerations. Therefore, based on the open-source nature of the model, KD of LLMs is frequently categorized into white-box KD \cite{gu2023knowledge, agarwal2024onpolicy, wen2023fdivergence, DBLP:conf/icml/KoKCY24, DBLP:journals/corr/abs-2410-13944} for open-source LLMs \cite{touvron2023llama, yang2023baichuan, jiang2024mixtral} and black-box KD \cite{zhou2023bridging, chen2024knowledge} for closed-source LLMs \cite{brown2020language, openai2023gpt4}. 

In this work, we focus on white-box KD since the findings can be more applicable. 
The process of white-box KD is similar to traditional KD, often utilizing a teacher-student framework to learn the rich probability distribution of the teacher model through soft labels.

\subsection{White-Box KD Objectives}

KD based on KL works well in previous models and tasks. For prediction of individual tokens, the formula is expressed as: $D_{\text{KL}}(P \,||\, Q) = \sum_i P(i) \log(P(i)/Q(i))$, where $P$ and $Q$ are predicted distributions by the teacher and student models, respectively. Then, \newcite{kim2016sequencelevel} propose SeqKD, wherein the results of the teacher model's beam search are employed as an approximation for the sequence-level KL. To minimize the KL divergence, $Q(i)$ needs to be as large as possible when $P(i)$ is large, but the value of $Q(i)$ has little impact when $P(i)$ is small. Therefore, $Q(i)$ is likely to be assigned a disproportionately high probability value when $P(i)$ is very small, as shown in Figure \hyperref[Fig.kl]{1}.

To make the student model pay more attention to peak predictions, more studies \cite{gu2023knowledge, agarwal2024onpolicy} use RKL as the distillation objective, expressed by the formula: $D_{\text{RKL}}(P \,||\, Q) = \sum_i Q(i) \log(Q(i)/P(i))$. RKL ensures that $Q(i)$ is not assigned an unreasonably high probability when $P(i)$ is small. However, when $P(i)$ is large, both large and small values of $Q(i)$ result in a low $D_{\text{RKL}}$ value. This can lead to the student model missing some peaks in a multi-modal distribution during learning, as shown in Figure \hyperref[Fig.kl]{1}.

To avoid the mode problems with KL and RKL, \newcite{wen2023fdivergence} introduce symmetric divergence functions to seek a balance between these two extremes, such as Jensen-Shannon Divergence (JSD) and Total Variation Distance (TVD). They also extend these word-level objectives to sequence-level.

Unlike the above, \newcite{DBLP:journals/corr/abs-2404-02657} argue that the predicted distribution of LLMs does not meet the conditions of continuity and standard Gaussian distribution. They theoretically and practically demonstrate that KL and RKL actually share the same optimization objective in LLMs' KD. Additionally, they point out that KL and RKL follow different optimization paths, with one fitting from the head part first and the other from the tail part first, and propose dynamically combining KL and RKL into Adaptive Kullback-Leiber (AKL).

Inspired by existing research, we aim at exploring a KD objective that yields better multi-modal learning ability. However, we first need to validate the following two questions: 

\begin{itemize}
    \item [1.]
    \textit{Does enhancing the learning of multi-modal distributions benefit the performance of student models? }
    \item [2.]
    \textit{Can existing distillation objectives ensure student models learn well from multi-modal distributions in KD of LLMs?}
\end{itemize}


\begin{figure*}[ht] 
\centering 
\includegraphics[width=0.40\textwidth]{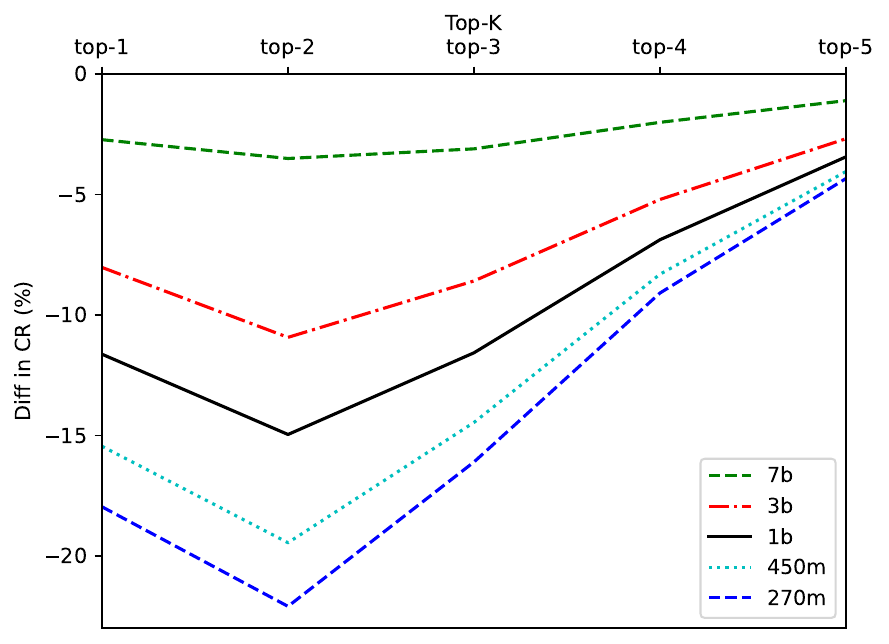} 
\includegraphics[width=0.47\textwidth]{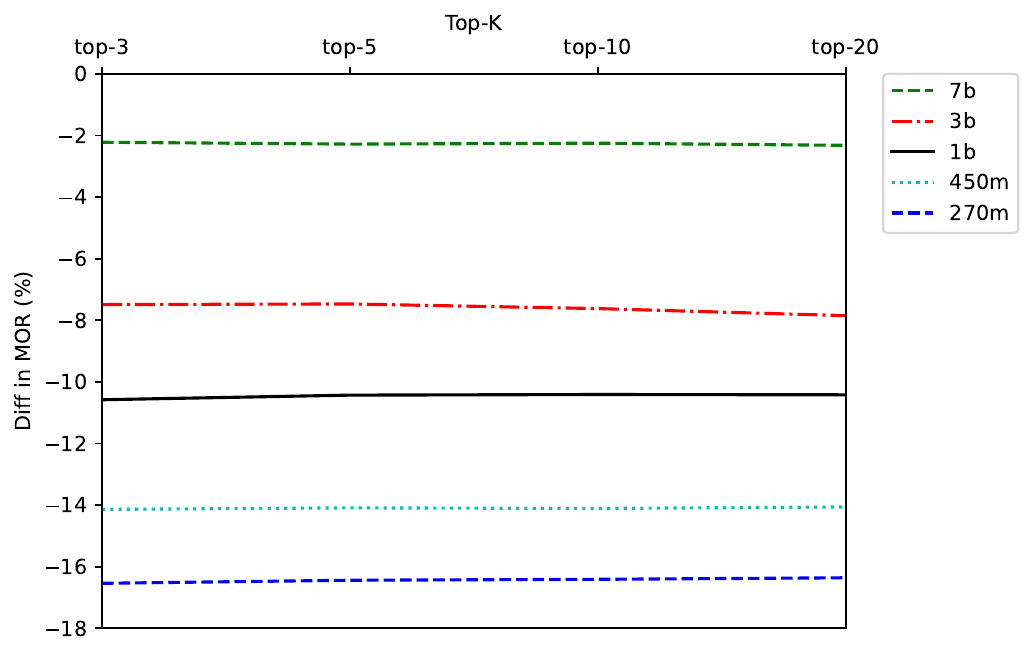} 
\caption{The degree of consistency between different models and the peak predictions of the 70B model. The horizontal axis represents the range of top-k predictions. For better presentation, we set the vertical axis as the difference between the CR or MOR of the current model and the corresponding results of the 13B model.} 
\vspace{-0.2cm}
\label{Fig2} 
\end{figure*}

\section{Preliminary}

\subsection{Metrics}

The purpose of enhancing the ability to learn multi-modal distributions is to make the peak predictions of student and teacher models closer. Therefore, we first need to define the criteria for evaluating the similarity of the peak predictions of two models. Since top-k sampling is the most common sampling strategy in language models, and peak predictions well correspond to the results of top-k sampling. Hence we convert the measurement of the consistency of multi-modal predictions into the measurement of the consistency of top-k sampling results.

In this paper, we introduce two metrics to assess the consistency of top-k sampling results. We use the consistency rate (CR) and the mean overlap rate (MOR) of the predicted top-k samples on the test set to evaluate how similar the peak predictions of the two models are. Specifically, CR measures the percentage of cases where the two models make identical top-k predictions, including both the categories and their order. The overlap rate (OR) measures the proportion of shared categories in the top-k predictions of both models, ignoring the order and position. MOR is the average OR across the whole test set.

\subsection{Motivation for Multi-Modal Distribution Learning}
\label{sec3_2}

We believe that peak predictions reflects the performance of language models, not just top-one prediction. Through experiments in this section, we demonstrate that the quality of peak predictions and the model's capabilities are highly correlated, thereby demonstrating the value of learning multi-modal distributions in the enhanced KD process.

For experiments, we select several models with significant performance gaps and a shared vocabulary. These models come from two families: Llama-2 released by \newcite{touvron2023llama} and OpenELM released by \newcite{DBLP:journals/corr/abs-2404-14619}. We use models with parameter sizes of 270M, 450M, 1B, 3B, 7B, 13B and 70B. Clearly, within the above models, models with larger parameter sizes exhibit stronger performance.

Among them, the 70B model has significantly more parameters and better performance compared to the other models, thus we can consider it as the ground truth. We use CR and MOR on test set to evaluate how closely the peak predictions of the other models align with those of the 70B model. This allows us to verify the relationship between model performance and the quality of peak predictions.

For the test set, we sample 5,000 slices from SlimPajama \cite{cerebras2023slimpajama}. The visualised experimental results are in Figure \hyperref[Fig2]{2}, and we also show the specific numerical results in Appendix \hyperref[sec:appendix]{B}.

\textbf{Analysis} The experimental results indicate that the closer a model's peak prediction consistency is to the strongest model, the better its performance. Thus, there is a direct correlation between the quality of peak predictions and model performance, not just the top-one prediction. Therefore, enhancing learning about multi-modal distributions is crucial during the KD process of LLMs.

\subsection{Validation for Existing KD Objectives}
\label{3-3}

We evaluate existing distillation objectives through experiments to determine if they enable student models to learn the multi-modal distributions of teacher models effectively during the KD process.

We respectively verify the similarity between the student model after distillation training and the teacher model's peak predictions under different distillation objectives to judge their learning performance on multi-modal distributions. Similarly, we evaluate the top-k prediction results over multiple ranges using CR and MOR. Since this paper focuses on distillation objectives for soft labels, the loss in all distillation experiments in this paper only includes soft targets.

To enhance the validity of the conclusion, we also conduct verification in real scenarios. We use Llama-2-7B \cite{touvron2023llama} and TinyLlama-1.1B \cite{zhang2024tinyllama} as teacher and student models, SlimPajama as the train set. Similar to the conventional settings, we use a learning rate and batch size that align with the practical pre-training task. Since we are assessing learning ability, we validate on the training data that has already been learned to evaluate the extent to which the student model's peak predictions after KD matches the teacher model. In particular, to make the differences in the multi-modal distribution learning ability of different distillation objectives more convincing, we increase the number of training epochs to 20.

\begin{figure*}[ht] 
\centering 
\includegraphics[width=0.43\textwidth]{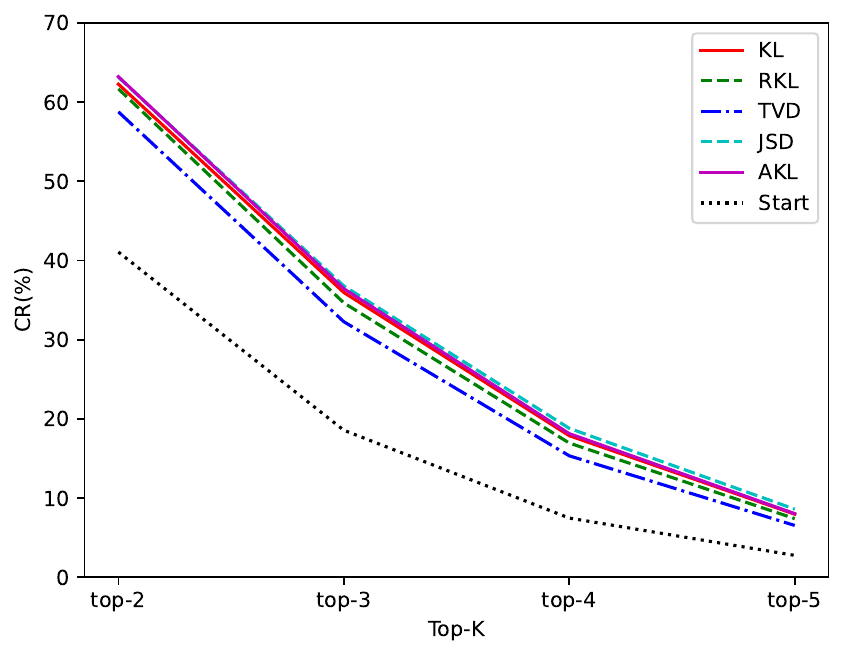} 
\includegraphics[width=0.43\textwidth]{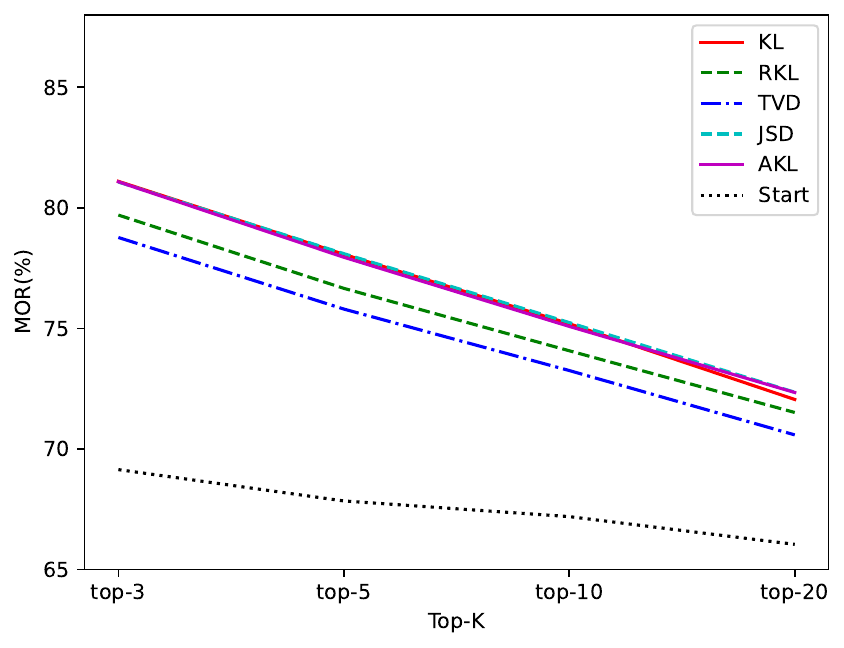} 
\vspace{-0.1cm}
\caption{Degree of agreement between student model and teacher model peak predictions after 20 epochs under existing KD objectives.} 
\vspace{-0.2cm}
\label{Fig3} 
\end{figure*}

The experimental results are shown in Figure \hyperref[Fig3]{3}, and more specific experimental setups and numerical metrics can be found in Section \hyperref[5-3]{5.3}.

\textbf{Analysis} Based on the results in Figure \hyperref[Fig3]{3}, existing distillation objectives show no significant disparity in their impact on the ability to learn multi-modal distributions. This further confirms \citeposs{DBLP:journals/corr/abs-2404-02657} view that KL and RKL share the same optimization objective in KD of LLMs. But more importantly, even after 20 epochs of training, student models still exhibit deficiencies in learning multi-modal distributions under the existing distillation objectives. Therefore, further exploration is necessary to identify distillation metrics that can enhance the model's capability to learn multi-modal distributions.

\section{Method}
\label{sec4}

Existing distillation objectives bring the two distributions closer by minimizing the distance between the teacher's and the student's predicted distributions. Although these distillation objectives can align the student's predicted distribution with the teacher's after a sufficient number of steps in theory, their efficiency in learning multi-modal distributions in practical scenarios still needs further improvement. Consequently, we aim to introduce additional optimization objectives to enhance the learning of peak predictions.

\begin{figure}[h] 
\centering 

\includegraphics[width=0.45\textwidth]{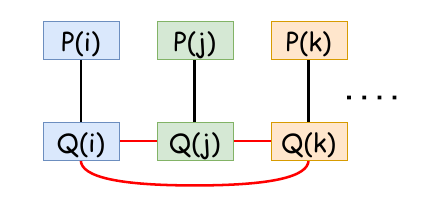} 
\vspace{-0.1cm}
\caption{Comparison of computational objects on peak predictions. The black lines represent existing distillation objectives and the red lines represent our method.} 
\label{Fig4}
\vspace{-0.2cm}
\end{figure}

The direct optimization objective of the existing distillation objectives is the distance between two distributions. The methods for calculating the distance of these distillation objectives differ, but they compute the same objects, as $\mathcal{L}_{\text{logits}}=\sum_i distance(P(i),Q(i))$. Therefore, existing distillation objectives only calculate the distance between each individual category, without utilizing the relationship among categories, as the black lines shown in Figure \hyperref[Fig4]{4}.

In this work, we enhance the learning of peak predictions in KD of LLMs by introducing a new optimization objective of \textit{word-level ranking loss}. The new optimization objective focuses on the prediction order of high probabilities between student and teacher models, enabling the student model to match the teacher model on critical predicted categories.

Our specific approach focuses on the top-$k$ predicted tokens from both the teacher and student models. We calculate the consistency by comparing the probability order of these tokens in the teacher model with the probability order in the student model. This method straightforwardly enhances the consistency of top-$k$ predictions between two multi-modal distributions, thereby strengthening the alignment of peak predictions between the student and teacher models. Importantly, the computational objects of our ranking loss are the probability values in the prediction sequences of the union of the teacher's and student's top-k predictions, not just the teacher's top-k predictions. This ensures that the excessively high predictive probability in the student model are also reasonably optimized.

Our approach allows that during the optimization process, the calculation of peak predictions is not limited to comparisons within a single category. As the red lines shown in Figure \hyperref[Fig4]{4}, $Q(i)$ needs to be compared with $Q(j)$ and $Q(k)$ based on the ranking position of $P(i)$ in the teacher's predictions to minimize the ranking loss.

We consider Spearman's rank correlation coefficient (SRCC) as the target for the measurement of ranking consistency. Compared to the Pearson coefficient, which also measures order consistency, SRCC only considers the consistency in the order of two sets of arrangements, without taking into account the correlation of the actual element values. We prefer that the ranking loss focuses more on the consistency of the predicted categories and probability values are non-linear relationships, therefore we select SRCC as the optimization objective for ranking loss, as
\begin{equation}
\mathcal{L}_{\text{Ranking}} = 1 - \rho_{\text{srcc}}(p, q) = 1 - \frac{\text{Cov}(R_p, R_q)}{\sigma_{R_p} \cdot \sigma_{R_q}}
\end{equation}
where $p$ and $q$ are subsets of distributions $P$ and $Q$, respectively, and each subset represents the probability values on the respective distributions for the union of top-$k$ predictions. $R_p$ denotes the rank index of $p$, $\sigma_{R_p}$ is the standard deviation of $R_p$, and $\text{Cov}(R_p, R_q)$ is the covariance of $R_p$ and $R_q$. 

Although sorting operations are theoretically non-differentiable, existing work \cite{blondel2020fast, ramzi2023optimization} has implemented differentiable ranking operator suitable for stochastic gradient descent. Several studies \cite{huang2022relational, rudd2022efficient, wang2023monoskd} have used SRCC as an optimization objective in other research areas based on such operators.

Overall, our method fully utilizes the peak predictions information from both the teacher and student models. Compared to previous methods that calculate loss within a single category, our approach further optimizes using probability values between categories. As shown in Figure \hyperref[Fig4]{4}, when combined with existing objectives, the fused objective allows the student model to more comprehensively learn the peak predictions of the teacher model from two different perspectives, showing excellent compatibility.

\section{Experiments}

In this section, we verify the effectiveness of our method on the pre-training and downstream tasks.

\subsection{Baselines}

To validate the effect of ranking loss, we introduce some distillation objectives that also focus on soft label learning as baseline methods.


\textbf{Supervised Fine-Tuning(SFT)} We verify the effectiveness of KD by comparing with direct fine-tuning.

\textbf{Word-Level Distillation}  We choose four word-level distillation objectives that are used more frequently in recent work: \textbf{KL}, \textbf{RKL}, \textbf{JSD} and \textbf{TVD}. Afterwards, we validate the boosting effect of ranking loss when combined with these base distillation objectives.

\textbf{SeqKD} \cite{kim2016sequencelevel}
This method is representative of sequence-level distillation, which approximates sentence-level KL as fine-tuning on teacher-generated data.

\textbf{$f$-DISTILL} \cite{wen2023fdivergence} We compare with the sequence-level KL in $f$-DISTILL (abbreviated as FD). Similar to SeqKD, FD also relies on teacher-generated data, but adopts soft labels for training. In particular, we have not compared with other methods in $f$-DISTILL because they rely on sampling directly from the student model, which is not as effective without pre-distillation \cite{shleifer2020pretrained}.

\textbf{Adaptive Kullback-Leiber divergence} (\textbf{AKL}, \citealt{DBLP:journals/corr/abs-2404-02657}) For the calculation process of AKL, we use the same experimental setup as in the original paper. We set the hyperparameter $\mu$ as 0.5 and the gap function $\epsilon(p(z),q(z))=|p(z)-q(z)|$.

\subsection{Datasets and Models}

Dataset used in the pre-training task:

\textbf{SlimPajama} \cite{cerebras2023slimpajama} A high-quality pre-training dataset with a mixture of data in reasonable proportions. We test the CR and MOR on the training set to assess how efficiently the student model learns the multi-modal predictive distribution.

Datasets used in downstream tasks:

\textbf{GSM8K} \cite{cobbe2021gsm8k} A high-quality mathematical reasoning dataset, each entry has a complete reasoning process, making it very suitable for KD tasks. It contains 8.5k challenging grade school math word problems. We follow the dataset's original test set division, with 1,319 samples as the test set and the rest as the training set. We use answer accuracy as the evaluation metric.

\textbf{databricks-dolly-15k} \cite{DatabricksBlog2023DollyV2} A directive fine-tuning dataset covering various tasks. We randomly select 14,000 samples for the training set and 800 samples for the test set. We use ROUGE scores \cite{lin-2004-rouge} as the evaluation metric to test the generative performance.

\textbf{Xsum} \cite{Narayan2018DontGM} An extensively used text summarization dataset. We randomly select 20,000 samples for the training set and 1,000 samples for the test set. Evaluation is also conducted through ROUGE scores.

For all KD tasks, we employ Llama-2-7B \cite{touvron2023llama} as the teacher model and Tinyllama-1.1B \cite{zhang2024tinyllama} as the student model. Prior to distillation, we have fine-tuned the teacher models on the respective datasets to adapt it to the tasks. Except in the GSM8K task, we directly use gsm8k-rft-llama7b2-u13b model released by \newcite{yuan2023scaling} due to its excellent performance.

More details can be found in Appendix \hyperref[sec:appendix]{B}.

\subsection{Results in the Pre-Training Task}
\label{5-3}

\begin{table*}[h]
  \small
  \centering
  \begin{center}
    \begin{tabular}{|l|c|ccccc|cccc|}
    \hline
      \multirow{2}{*}{{\textbf{Loss}}} & \multirow{2}{*}{\textbf{Perplexity$\downarrow$}} & \multicolumn{5}{c|}{\textbf{\makecell{CR$\uparrow$ (\%)}}} & \multicolumn{4}{c|}{\textbf{\makecell{MOR$\uparrow$ (\%)}}} \\
      \cline{3-11}
      & & top1 & top2 & top3 & top4 & top5 & top3 & top5 & top10 & top20 \\
      \hline
      Start & 10.83 & 75.52 & 41.17 & 18.59 & 7.44 & 2.76 & 69.14 & 67.84 & 67.19 & 66.05 \\
      \hline
      KL & 7.85 & 89.44 & 62.27 & 35.96 & 17.88 & 7.97 & 81.10 & 78.07 & 75.19 & 72.05\\
      \textbf{KL+R} & \textbf{7.81} & \textbf{90.59} & \textbf{69.08} & 44.54 & 25.21 & 12.65 & 86.00 & 85.04 & 83.39 & 76.75\\
      \hline
      RKL & 8.25 & 90.03 & 61.67 & 34.65 & 16.95 & 7.41 & 79.70 & 76.66 & 74.07 & 71.51\\
      \textbf{RKL+R} & 7.99 & 90.29 & 67.27 & 41.94 & 22.74 & 10.99 & 84.75 & 83.56 & 81.76 & 75.50\\
      \hline
      JSD & 8.23 & 89.98 & 63.14 & 36.79 & 18.81 & 8.60 & 81.07 & 78.10 & 75.24 & 72.35\\
      \textbf{JSD+R} & 8.01 & 90.15 & 69.03 & \textbf{45.48} & \textbf{26.46} & \textbf{13.70} & \textbf{86.60} & \textbf{86.02} & \textbf{84.79} & \textbf{76.94}\\
      \hline
      TVD & 8.54 & 88.66 & 58.77 & 32.27 & 15.34 & 6.55 & 78.77 & 75.80 & 73.25 & 70.58\\
      \textbf{TVD+R} & 7.92 & 89.87 & 67.65 & 43.21 & 24.09 & 12.68 & 85.62 & 84.89 & 83.43 & 76.34\\
      \hline
      AKL & 7.93 & 90.36 & 63.21 & 36.44 & 18.14 & 8.04 & 81.08 & 77.95 & 75.08 & 72.34\\
      \textbf{AKL+R} & 7.86 & 90.50 & 68.49 & 43.64 & 24.39 & 12.14 & 85.60 & 84.54 & 82.69 & 76.31\\
      \hline
    \end{tabular}
    \caption{Learning situation of multi-modal distribution for data already learned in the pre-training task. "+R" represents that we have added an additional fixed-ratio ranking loss.}
    \label{tab1}
  \end{center}
  \vspace{-0.2cm}
\end{table*}

\begin{figure*} 
\centering 
\includegraphics[width=0.42\textwidth]{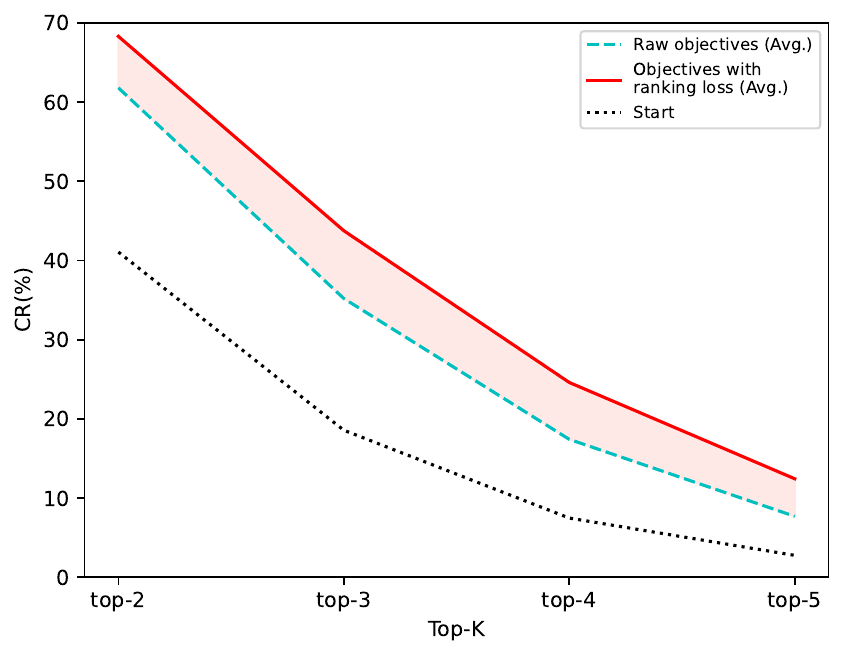} 
\includegraphics[width=0.42\textwidth]{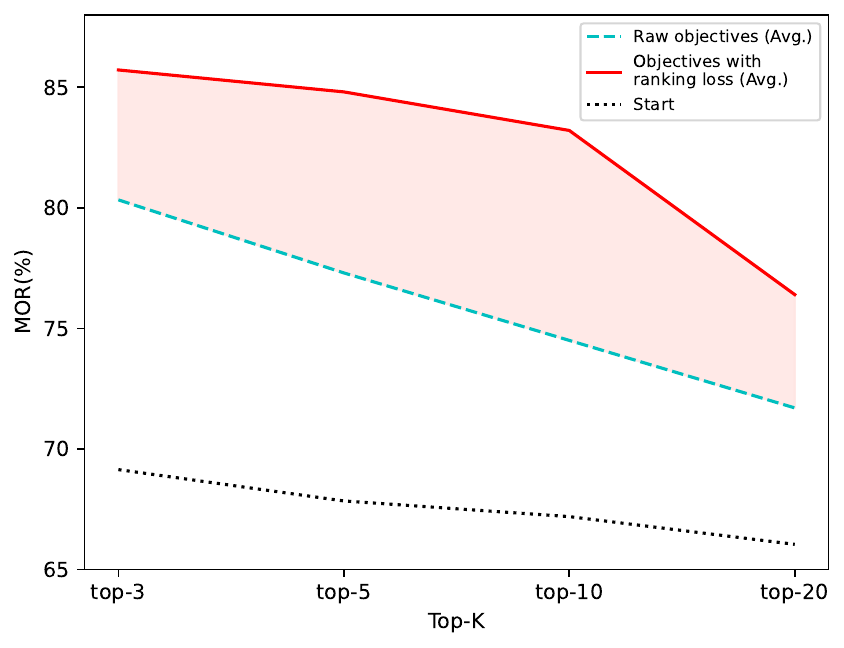} 
\vspace{-0.2cm}
\caption{Improvement in the learning ability of multi-modal distributions for existing distillation objectives by introducing ranking loss in the pre-training task. We average the results of the five distillation objectives before and after adding the ranking loss. The red area indicates the improved parts.} 
\vspace{-0.2cm}
\label{Fig5} 
\end{figure*}

\begin{table*}[ht]
  \small
  \begin{center}
    \begin{tabular}{|l|cc|ccc|ccc|}
    \hline
      \multirow{2}{*}{{\textbf{Method}}} & \multicolumn{2}{c|}{\textbf{\makecell{GSM8K}}} & \multicolumn{3}{c|}{\textbf{\makecell{Dolly}}} & \multicolumn{3}{c|}{\textbf{\makecell{Xsum}}} \\
      \cline{2-9}
      & \textbf{Correct\_Num} & \textbf{Score} & \textbf{R-1} & \textbf{R-2} & \textbf{R-L} & \textbf{R-1} & \textbf{R-2} & \textbf{R-L}\\
      \hline
      Teacher & 682 & 51.71 & 33.55 & 17.70 & 31.55 & 41.18 & 18.81 & 34.39\\
      \hline
      SFT & 227 & 17.21 & 23.60 & 10.18 & 22.29 & 33.27 & 11.84 & 26.57\\
      \hline
      SeqKD & 200 & 15.16 & 22.73 & 9.64 & 21.20 & 34.78 & 13.20 & 28.64\\
      \hline
      \textbf{Rank-5} & 229 & 17.36 & 25.74 & 12.39 & 23.77 & 34.71 & 12.53 & 28.18\\
      \textbf{Rank-15} & 236 & 17.89 & 26.13 & 12.27 & 24.16 & 34.93 & 12.55 & 28.25\\
      \hline
      KL & 219 & 16.60 & 23.09 & 10.12 & 21.82 & 34.62 & 13.06 & 28.24\\
      \textbf{KL+R} & \textbf{267}(+48) & \textbf{20.24} & 24.72 & 11.47 & 23.44(+1.62) & 35.41 & \textbf{13.61} & 28.93(+0.69)\\
      \hline
      RKL & 132 & 10.01 & 23.84 & 10.19 & 22.55 & 32.97 & 11.41 & 26.72\\
      \textbf{RKL+R} & 191(+59) & 14.48 & 26.27 & 12.44 & 24.40(+1.85) & 35.20 & 12.83 & 28.60\textbf{(+1.88)}\\
      \hline
      JSD & 160 & 12.13 & 25.41 & 11.50 & 23.67 & 35.13 & 13.33 & 28.51\\
      \textbf{JSD+R} & 227(+67) & 17.21 & 26.60 & 12.62 & 24.81(+1.14) & 35.18 & 13.18 & 28.71(+0.20)\\
      \hline
      TVD & 0 & 0.00 & 26.21 & 12.25 & 24.68 & 34.84 & 12.74 & 28.45\\
      \textbf{TVD+R} & 240\textbf{(+240)} & 18.20 & \textbf{27.28} & 13.44 & \textbf{25.35}(+0.67) & \textbf{35.84} & 13.59 & \textbf{29.40}(+0.95)\\
      \hline
      FD & 194 & 14.71 & 23.29 & 9.64 & 21.91 & 34.27 & 12.69 & 27.94\\
      \textbf{FD+R} & 265(+71) & 20.09 & 25.19 & 11.66 & 23.40(+1.49) & 35.37 & 13.17 & 28.88(+0.94)\\
      \hline
      AKL & 215 & 16.30 & 24.59 & 10.56 & 23.20 & 34.19 & 12.79 & 27.96\\
      \textbf{AKL+R} & 235(+20) & 17.82 & 26.86 & \textbf{13.51} & 25.16\textbf{(+1.96)} & 35.09 & 13.34 & 28.68(+0.72)\\
      \hline
    \end{tabular}
    \caption{Experimental results on test set of downstream tasks. "+R" represents that we have added an additional fixed-ratio ranking loss. "R-1", "R-2", and "R-L" are abbreviations for ROUGE-1, ROUGE-2, and ROUGE-L, respectively. We have also marked the improvement after combining the original objectives with ranking loss in parentheses for the most important metrics in the table.}
    \label{tab2}
  \end{center}
\end{table*}




In pre-training task, we investigate the impact of introducing ranking loss on improving the alignment of top-k predictions between student and teacher models. The reason for the validation on pre-training task rather than downstream tasks is that student model can be easier to capture task-specific peak predictions patterns on a single task, thereby approaching the distribution of the teacher model more closely. Inversely, the richness of categories and types in pre-training data makes it more likely that the proximity between the student and teacher model predictions is a result of KD training.

\textbf{Training} With the same experimental setup as in Section \hyperref[3-3]{3.3}, we test the effectiveness of these five distillation objectives combined with ranking loss. For ranking loss, we align predictions of the top 15 between student and teacher models. We employ a fixed ratio for loss allocation, as $\mathcal{L}_{\text{total}} = 2 \cdot \mathcal{L}_{\text{Ranking}} + \mathcal{L}_{\text{logits}}$. The choice of ranking ranges and the allocation of losses is discussed in Appendix \hyperref[C]{D}. For the convenience of subsequent ablation analysis, in addition to the multi-modal distribution similarity, we also evaluate the perplexity and CR of the top-1 prediction. The experimental results are presented in Table \hyperref[tab1]{1}, while more detailed training information can be found in Appendix \hyperref[sec:appendix]{B}. We also present extra experiments in Appendix \hyperref[ood]{E} to verify the generalization ability of our method.

\textbf{Results} Table \hyperref[tab1]{1} shows the various metrics measured on the learned data for the student model before training and after KD training with different objectives. The results indicate that when combined with ranking loss, all five different objectives significantly improve the student model in terms of the multi-modal consistency metric with the teacher model during the distillation process. Moreover, while the similarity of the multi-modal distribution improves, the top-1 accuracy and perplexity performance are not negatively affected and even show slight improvements. To make the metrics for the similarity of multi-modal distributions more intuitive, we present visualized results in Figure \hyperref[Fig5]{5}, which show the improvement of ranking loss in a clearer manner. Compared to the mean scores of original objectives, our method improves the CR metric by approximately 30\% to 95\% across different ranges, and the OR metric is improved by about 50\% to 120\% across different ranges.

Overall, our approach significantly improves the efficiency of aligning multi-modal predictions between the student and teacher models during the distillation process. Furthermore, for the five different distillation objectives, ranking loss demonstrates stable and effective improvements, showcasing its excellent compatibility with all these commonly used KD objectives.

\subsection{Results in Downstream Tasks}
\label{5-4}

Although experiments in pre-training task fully validate that our method achieves the motivation of improving multi-modal prediction distribution learning during the KD process, its effectiveness in improving the performance of the student model after distillation still requires further verification. 
Therefore, we conduct thorough validation of our method on datasets from multiple different downstream tasks to demonstrate its contribution to improving the performance of the student model.

\textbf{Training}  We conduct experimental verification on all baselines and downstream task datasets, and introduce ranking loss on various baselines containing word-level distillation objectives. Specifically, we also conduct experiments with only ranking loss to evaluate the impact of peak alignment on downstream task effectiveness, as "Rank-$k$" to align top-$k$ predictions. For fused loss, we align predictions of top-5 between teacher and student models to enhance applicability on downstream tasks. We use the same loss allocation ratio as the pre-training task. We also discuss the choice of ranking ranges and the allocation of losses in detail in Appendix \hyperref[C]{D}, including fixed-rate and dynamic-rate losses allocations. The experimental results are presented in Table \hyperref[tab2]{2}, more detailed training information can be found in Appendix \hyperref[sec:appendix]{B}.

\textbf{Results} The experimental results show the scores of the student models after training with the baseline method and our method, where the highest scores for each metric of each task are achieved by our method. The table also shows the improvements after combining our proposed ranking loss with existing word-level distillation objectives. This combination enhances the performance of existing methods on nearly all metrics, with significant improvements on most. Specifically, after introducing the ranking loss, most of the accuracy improvements of the student model on the GSM8K test set are over 20\% compared to original objective, most ROUGE-L scores on the Dolly test set improves by over 1.0 point, and most ROUGE-L score on the Xsum test set improves by over 0.7 points. Especially, when only using the ranking loss, our method learns only the peak predictions, which account for only about 0.0001\% of the total categories, yet surpasses most existing distillation objectives in evaluations across multiple tasks. This not only demonstrates the importance of learning peak predictions, but also showcases the outstanding performance of our method in peak predictions alignment.

In summary, experiments on downstream tasks validate that our method significantly improves the performance of student model in the KD process. Compared to other optimization objectives of soft labels, our method demonstrates excellent competitiveness and compatibility. In addition, the consistent performance improvements across different datasets confirm the generality and robustness of our approach.

\section{Further Analysis}

\subsection{Ablation Study}

In this section, we use further ablation analysis to reveal whether the performance improvement brought by our method is due to the enhanced ability of multi-modal distribution learning during the KD process.

Based on the respective results and analysis in Section \hyperref[5-3]{5.3} and \hyperref[5-4]{5.4}, we can conclude a preliminary ablation conclusion that the improvement in the ability to learn multi-modal distributions and the enhancement in downstream task performance are indeed attributed to the introduction of ranking loss.

Furthermore, we can observe in Table \hyperref[tab1]{1} that although the improvement of the top prediction accuracy is modest by introducing the ranking loss, the peak prediction alignment at other positions is significantly improved. Therefore, ranking loss does not have a significant impact on the actual learning efficiency of the top-1 prediction.

However, in downstream tasks, we use a greedy decoding strategy, which should not exhibit better performance when there is no significant improvement in the consistency of the top prediction. In fact, the reason is that in downstream tasks, we make the student model learn task-related prediction patterns more efficiently by enhancing its ability to learn from multi-modal distributions. Therefore, our method achieve better performance with the same number of training steps.

Based on the above analysis, we can more confidently get the conclusion that ranking loss primarily improves the fitting of multi-modal distributions, with modest impact on the alignment of top-one predictions. Therefore, the existing objectives without the addition of ranking loss can be regarded as an ablation of the ability to learn multi-modal distributions, leading to worse results. This further shows the importance of aligning multi-modal predictive distributions in KD of LLMs.

\subsection{Case Study}

Based on the results in Table \hyperref[tab2]{2}, under our experimental setup, the accuracy of TVD on GSM8K is 0. In fact, this is due to TVD's inadequacy in peak predictions learning, which leads to its failure to grasp the answering norms of GSM8K. 

We conduct a case study within this interesting phenomenon in Appendix \hyperref[D]{G} to further analyze and demonstrate the importance of peak predictions learning.

\section{Conclusion}

In this paper, we propose ranking loss based knowledge distillation, a new objective function that improves the efficiency of aligning peak predictions of student and teacher models during white-box KD. We verify the importance of aligning multi-modal distributions through experiments and highlight the inefficiency of existing KD objectives in learning multi-modal distributions. Most importantly, we propose a word-level ranking loss to the existing KD objectives for more efficient alignment of multi-modal distributions. Our extensive experiments clearly demonstrate that our method effectively improves the multi-modal distribution alignment between teacher and student models, leading to significant performance gains in different downstream tasks.

\section*{Limitations}

Due to the extensive experiments have been conducted in this paper and the limitations of computational resources, we only perform distillation experiments on models within the Llama \cite{touvron2023llama} architecture. However, the existing generative models often have similar structures, and the Llama model family is one of the most widely used, making this study still highly applicable. We will also conduct experiments on other model families as future work.

Additionally, we encourage combining our proposed method with existing distillation objectives to achieve optimal performance. Although this introduces additional computation, this burden becomes negligible due to existing operators and our code optimization (only adding about 1\% extra training time). We show the time consumption in Appendix \hyperref[B]{C}.

\section*{Acknowledgments}

This work is supported by National Key R\&D Program of China 2022ZD0160602 and the Natural Science Foundation of China 62122088.

\bibliography{custom}

\appendix

\section{Proportion of Multi-Modal Distributions}
\label{A}

Obviously, the diversity of natural language leads to the prediction of language models exhibiting a multi-modal characteristic. In this section, we quantitatively demonstrate through experiments the proportion of the multi-modal distribution in the overall prediction distribution, further proving the necessity of enhancing the ability to learn multi-modal distributions during the KD process.

\begin{figure} 
\centering 
\subfigure[$temperature=1.0$]{
\includegraphics[width=0.49\textwidth]{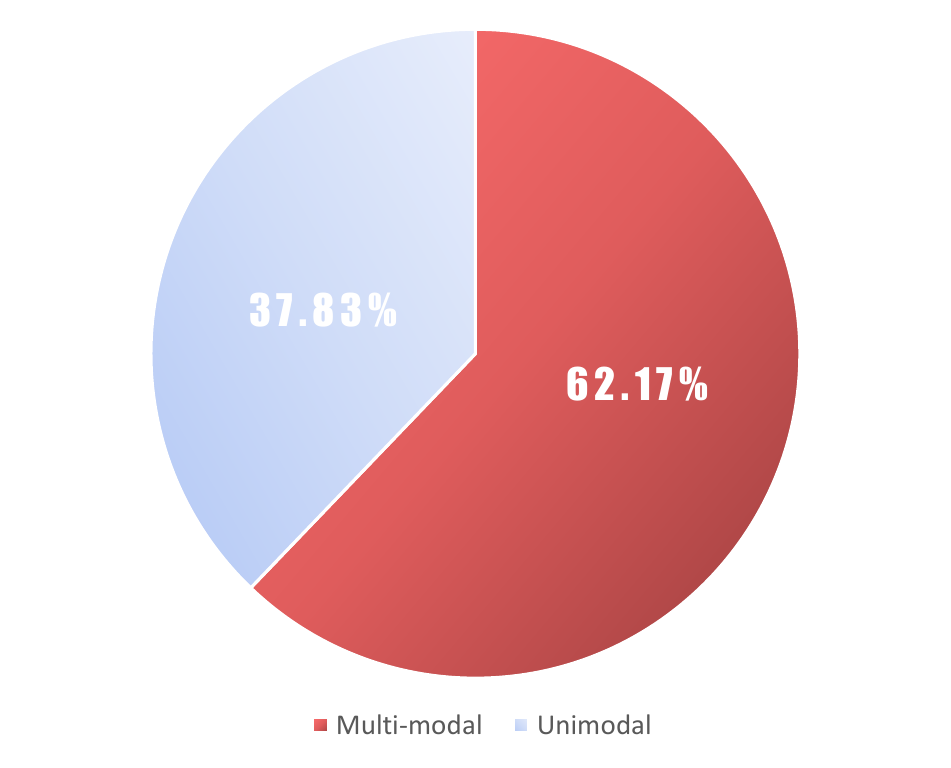} 
}
\subfigure[$temperature=0.8$]{
\includegraphics[width=0.49\textwidth]{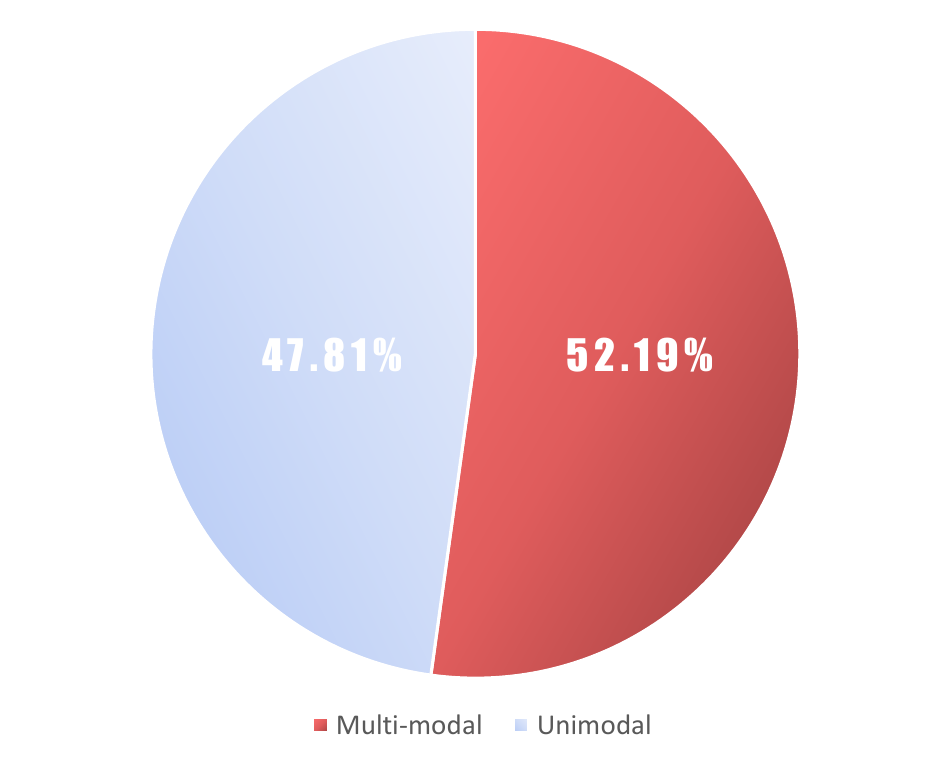} 
}
\subfigure[$temperature=0.6$]{
\includegraphics[width=0.49\textwidth]{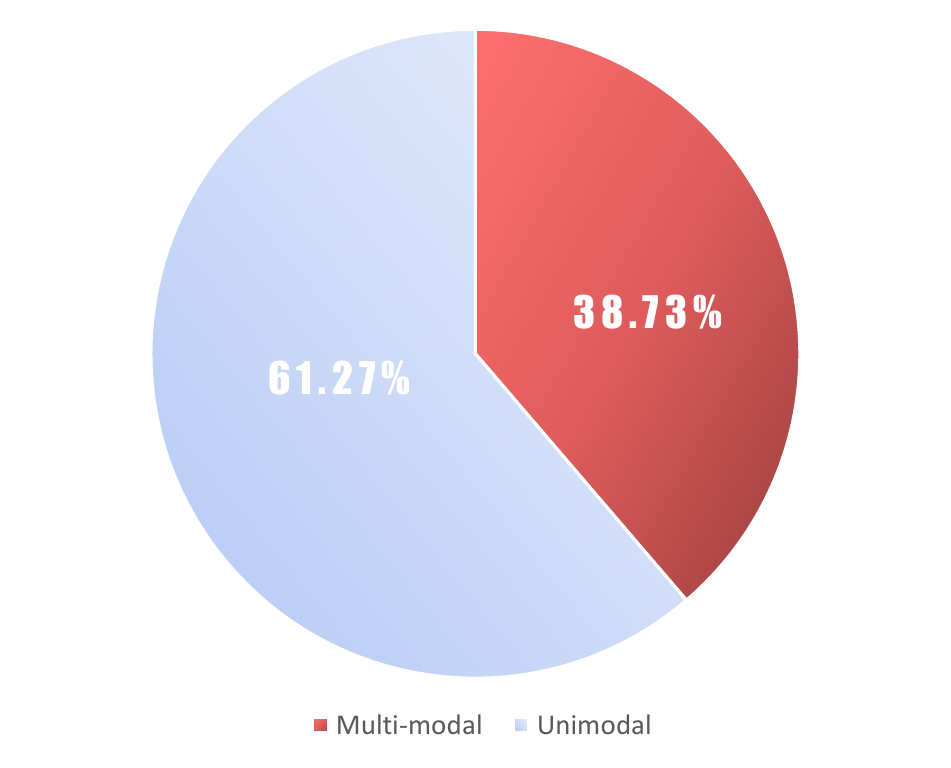} 
}
\caption{The results of the proportion of multi-modal predictions on the test set.} 
\label{percent} 
\end{figure}

Top-p sampling \cite{holtzman2020curious} combined with sampling temperature is a commonly adopted method for LLMs, ensuring both the reliability and diversity of sampling results. We verify the proportion of multi-modal predictions (i.e., cases where the number of sampled results for the next acceptable token is greater than one) in all predictions made by Llama-2-7B \cite{touvron2023llama} using this sampling method on 5,000 samples (containing approximately 3M tokens) from SlimPajama \cite{cerebras2023slimpajama}. We use the common top-p sampling setting of $p=0.9$ and test the results with several commonly used sampling temperatures, as shown in Figure \hyperref[percent]{6}.

According to the results in Figure \hyperref[percent]{6}, when the sampling temperature is high, most prediction distributions exhibit multi-modal characteristics. Even when the sampling temperature is low, multi-modal distributions still account for a significant portion. Therefore, it is essential to strengthen the student model's ability to learn from multi-modal distributions during the KD process.

\section{Details of Experiments}
\label{sec:appendix}

To enhance the reproducibility of our experiments, we use open-source models and datasets in our experiments, and we also detail the information of the experimental setup in this section. 


\textbf{Models} For specific information of models, we use Llama-2 (\citealt{touvron2023llama}, Meta license) model family, OpenELM (\citealt{DBLP:journals/corr/abs-2404-14619}, apple-sample-code-license) model family, TinyLlama-1.1B-intermediate-step-1431k-3T (\citealt{zhang2024tinyllama}, Apache License 2.0) and gsm8k-rft-llama7b2-u13b (\citealt{yuan2023scaling}) which is based on Llama-2-7B. Notably, we select a stronger model as the teacher model for GSM8K because the test of GSM8K requires completely accurate results, and a teacher model with a low accuracy rate can provide very limited guidance.

\textbf{Datasets} For specific information of datasets, we use SlimPajama (\citealt{cerebras2023slimpajama}, Apache License 2.0), GSM8K (\citealt{cobbe2021gsm8k}, MIT License), databricks-dolly-15k (\citealt{DatabricksBlog2023DollyV2}, CC BY-SA 3.0 license) and Xsum (\citealt{Narayan2018DontGM}, MIT License).

\textbf{Hardware Environment} All the experiments are conducted in two or four A100 GPUs with 80GB of VRAM each, and each individual experiment takes no more than 2 hours to complete. Based on our experimental observations, we believe that using either two GPUs with 40GB (via DeepSpeed, \citealt{10.1145/3394486.3406703}) each or a single GPU with 80GB of VRAM is also sufficient. 

\textbf{Training Parameters} For all training process in our experiments, we use the AdamW optimizer \cite{DBLP:conf/iclr/LoshchilovH19}. We set the learning rate to 2e-5 and gradient clipping threshold to 1.0. In the experiments involving KD, the distillation temperature are set to 1.0. For the pre-training task, we set the training epoch to 20, each epoch contains 25 steps, sequence length to 400, and each step contains 0.1M tokens. For downstream tasks, we set batch size to 64 and train 2 epochs on GSM8K which contains less data, 1 epoch for others. And the sequence length is set to 2048 in all downstream tasks. Due to the large volume of our experiments, it is not feasible to run multiple times for each individual experiment. But to ensure consistency and comparability, all comparative experiments in our study are conducted with the same random seed 72.

\textbf{Test Setting} We use greedy decoding strategy in the testing of all downstream tasks.

\textbf{Code Details} For the code implementation of ranking loss, we utilise the differentiable ranking operator of \textit{torchsort}\footnote{\href{https://github.com/teddykoker/torchsort}{https://github.com/teddykoker/torchsort}} library, which is a python implementation of the differentiable ranking method proposed by \newcite{blondel2020fast}. For the computation of the ROUGE score, we performed it through \textit{rouge}\footnote{\href{https://github.com/pltrdy/rouge}{https://github.com/pltrdy/rouge}}, a python library.

\textbf{Others} We show in Table \hyperref[tab3]{3} the numerical results of the experiments in Section \hyperref[sec3_2]{3.2}.

\begin{table*}[h]
  \centering
  \begin{center}
    \begin{tabular}{|l|ccccc|cccc|}
    \hline
      \multirow{2}{*}{{\textbf{Model Size}}} & \multicolumn{5}{c|}{\textbf{\makecell{CR$\uparrow$ (\%)}}} & \multicolumn{4}{c|}{\textbf{\makecell{MOR$\uparrow$ (\%)}}} \\
      \cline{2-10}
      & top1 & top2 & top3 & top4 & top5 & top3 & top5 & top10 & top20 \\
      \hline
      13B & \textbf{82.47} & \textbf{50.54} & \textbf{26.40} & \textbf{12.42} & \textbf{5.39} & \textbf{75.30} & \textbf{73.94} & \textbf{73.36} & \textbf{72.27} \\
      \hline
      7B & 79.74 & 47.02 & 23.29 & 10.40 & 4.28 & 73.08 & 71.66 & 71.11 & 69.95\\
      \hline
      3B & 74.44 & 39.60 & 17.81 & 7.22 & 2.70 & 67.81 & 66.47 & 65.74 & 64.43\\
      \hline
      1B & 70.84 & 35.57 & 14.83 & 5.54 & 1.95 & 64.72 & 63.51 & 62.95 & 61.85\\
      \hline
      450M & 67.02 & 31.08 & 11.95 & 4.11 & 1.34 & 61.11 & 59.85 & 59.25 & 58.21\\
      \hline
      270M & 64.51 & 28.44 & 10.30 & 3.33 & 1.05 & 58.76 & 57.50 & 56.95 & 55.91\\
      \hline
    \end{tabular}
    \caption{The numerical results of the experiments in Section \hyperref[sec3_2]{3.2}, showing the degree of agreement in peak predictions between the different models and the 70B model.}
    \vspace{-0.2cm}
    \label{tab3}
  \end{center}
\end{table*}

\begin{table}[ht]

\centering
\begin{tabular}{cc}
\hline
\textbf{Loss} & \textbf{Total Training Time (s)} \\
\hline
KL & 1424 \\
\textbf{KL+R} & 1441 (+1.19\%) \\
\textbf{Rank-15} & \textbf{1370 (-3.79\%)} \\
\hline
\end{tabular}
\caption{
The computation time for KD with different losses on GSM8K for 2 epochs with 2 A100 GPUs.
}
\vspace{-0.2cm}
\label{tab4}
\end{table}

\section{Analysis of Computational Efficiency}
\label{B}

We list the elapsed time for distillation training two epochs on GSM8K for some of the distillation objectives, as shown in Table \hyperref[tab4]{4}. Based on the results, we can find that the introduction of ranking loss has a very minimal impact on computational burden, which can be ignored. Additionally, when using only ranking loss, the computational efficiency is improved by eliminating the need for $softmax$ operation on output logits.

\section{Analysis of Hyperparameter}
\label{C}

There are two hyperparameters in our experiments, the optimised range of ranking loss and the proportion of the loss allocation.

For the range of ranking loss, we perform a number of experiments upfront to determine the value of the take that works best on the downstream KD tasks. Indeed, optimising the top predictions from 5 to 15 works well and is in line with our motivation for proposing ranking loss. For the pre-training task, due to the large number of calculations that need to be performed, we recommend setting range $k$ to 15 because a larger range makes the sorting operator of the differentiable more stable.

And for downstream tasks, we recommend taking the range $k$ to be 10-15 when using ranking loss alone, and $k$ to be 5 for mixing with other distillation objectives such as KL. Because when mixing losses, other distillation objectives can bring the two distributions closer together on a broader scale, and a smaller range $k$ helps the ranking loss focus more on peak prediction to achieve better performance. In addition, small changes to this range have a small effect on the results, and dynamic value of the range tends to create a computational burden during batch calculations, so we do not dynamically adjust this hyperparameter. We also present the results on GSM8K in Table \hyperref[kvalue]{5} after applying different $k$ values to partial distillation objectives, to further validate the above conclusions.

\begin{table*}[ht]
\centering
\begin{tabular}{cccccccccc}
\hline
\textbf{Value of k} & 5 & 10 & 15 & 20 & 25 & 30\\
\hline
\textbf{Rank-k} & 229 & 240 & 236 & \textbf{246} & 236 & 223 \\
\textbf{KL+R} & \textbf{267} & 264 & 243 & 240 & 237 & 229 \\
\textbf{RKL+R} & \textbf{191} & 171 & 168 & 176 & 151 & 159 \\
\textbf{TVD+R} & \textbf{240} & 236 & 216 & 228 & 233 & 222 \\
\hline
\end{tabular}

\caption{
The number of correct instances in the test set of GSM8K when using different ranking consistency computation objectives as the optimization target.
}

\label{kvalue}
\end{table*}

For the losses allocation method, we have found in experiments that the ratio of ranking loss to other distillation objectives is better when the ratio is 1 to 3. Besides, small changes have little effect on the KD effect, so we suggest to adopt this lossed allocation directly, as $\mathcal{L}_{\text{total}} = 2 \cdot \mathcal{L}_{\text{Ranking}} + \mathcal{L}_{\text{logits}}$. As shown in Figure \hyperref[Fig4]{4}, $\mathcal{L}_{\text{Ranking}}$ and $\mathcal{L}_{\text{logits}}$ align two distributions from different perspectives, thus we believe static allocation is appropriate.

Moreover, we also propose a dynamic allocation of losses for asymmetric KL and RKL. We employ OR of top-$k$ predictions from both the teacher and student models as an indicator of the understanding of current input by the student model. With this indicator, we are able to guide the focus of student learning towards peak predictions of teacher model, particularly when the gap between the peak predictions of the student model and the teacher model is excessive. And the optimization efforts will naturally gravitate towards refining global information when peak predictions have aligned. For each individual prediction, the mixed loss can be expressed by following formula, as
\begin{equation}
\mathcal{L}_{\text{total}} = 2 \cdot \mathcal{L}_{\text{Ranking}} + \frac{|p^{k} \cap q^{k}|}{k} \cdot \mathcal{L}_{\text{logits}}
\end{equation}
where $p^{k}$ is the index set of the top-$k$ tokens in $P$, the value of $k$ is consistent with the range of ranking loss. Under the same experimental setup as in Section \hyperref[5-4]{5.4}, we show the effect of dynamic loss allocation on KL and RKL in Table \hyperref[dyna]{6}.

\begin{table*}[ht]
  \small
  \begin{center}
    \begin{tabular}{|l|cc|ccc|ccc|}
    \hline
      \multirow{2}{*}{{\textbf{Method}}} & \multicolumn{2}{c|}{\textbf{\makecell{GSM8K}}} & \multicolumn{3}{c|}{\textbf{\makecell{Dolly}}} & \multicolumn{3}{c|}{\textbf{\makecell{Xsum}}} \\
      \cline{2-9}
      & \textbf{Correct\_Num} & \textbf{Score} & \textbf{R-1} & \textbf{R-2} & \textbf{R-L} & \textbf{R-1} & \textbf{R-2} & \textbf{R-L}\\
      \hline
      Teacher & 682 & 51.71 & 33.55 & 17.70 & 31.55 & 41.18 & 18.81 & 34.39\\
      \hline
      KL & 219 & 16.60 & 23.09 & 10.12 & 21.82 & 34.62 & 13.06 & 28.24\\
      \textbf{KL+R} & 267(+48) & 20.24 & 24.72 & 11.47 & 23.44(+1.62) & 35.41 & 13.61 & 28.93(+0.69)\\
      \textbf{KL+R(Dynamic)} & \textbf{280}(+61) & \textbf{21.23} & 25.51 & 12.10 & 24.02\textbf{(+2.20)} & \textbf{35.51} & \textbf{13.62} & \textbf{29.10}(+0.86)\\
      \hline
      RKL & 132 & 10.01 & 23.84 & 10.19 & 22.55 & 32.97 & 11.41 & 26.72\\
      \textbf{RKL+R} & 191(+59) & 14.48 & 26.27 & \textbf{12.44} & \textbf{24.40}(+1.85) & 35.20 & 12.83 & 28.60(+1.88)\\
      \textbf{RKL+R(Dynamic)} & 204\textbf{(+72)} & 15.47 & \textbf{26.40} & 11.98 & 24.34(+1.79) & 35.46 & 13.40 & 29.04\textbf{(+2.32)}\\
      \hline
    \end{tabular}
    \caption{Comparison of dynamic allocation loss and fixed-rate allocation loss. "+R" represents that we have added an additional fixed-ratio ranking loss. "+R(Dynamic)" means that we have added an additional dynamic-ratio ranking loss.}
    \label{dyna}
  \end{center}
  \vspace{-0.4cm}
\end{table*}

Experimental results show that our dynamic allocation strategy achieves better scores on most tasks, further improving the effectiveness of distillation training. Other distillation objectives do not show head or tail bias, and in experiments it is found that fixed ratios of losses work better, so there is no need for dynamic losses allocation.

\section{Supplementary Experiments in the Pre-Training Task}
\label{ood}

We have shown the improvement in learning capability brought by our method in the main text. Additionally, we consider it necessary to show results on the test set data to prove the generalization ability of our method. Hence, we conduct extra experiments in this section to demonstrate the performance of our method on the test set.

\textbf{Training} We change the training steps to 2000, set the number of epochs to 1, and configure the batch size to approximately 0.5M tokens. Finally, we validate the performance on the out-of-distribution test dataset. The remaining training settings are consistent with those in Section \hyperref[3-3]{5.3}. The experimental results are presented in Table \hyperref[ood_pretrain]{7}.

\textbf{Results} The results in Table \hyperref[ood_pretrain]{7} show that our method also effectively improves the consistency of peak predictions between the student model and the teacher model on out-of-distribution data during the KD process. These results demonstrate the generalization ability of our method and complement the experimental results in the main text.

\begin{table*}[h]
  \small
  \centering
  \begin{center}
    \begin{tabular}{|l|c|ccccc|cccc|}
    \hline
      \multirow{2}{*}{{\textbf{Loss}}} & \multirow{2}{*}{\textbf{Perplexity$\downarrow$}} & \multicolumn{5}{c|}{\textbf{\makecell{CR$\uparrow$ (\%)}}} & \multicolumn{4}{c|}{\textbf{\makecell{MOR$\uparrow$ (\%)}}} \\
      \cline{3-11}
      & & top1 & top2 & top3 & top4 & top5 & top3 & top5 & top10 & top20 \\
      \hline
      Start & 10.93 & 75.37 & 41.07 & 18.57 & 7.48 & 2.77 & 69.14 & 67.84 & 67.19 & 66.04 \\
      \hline
      KL & 10.22 & 78.46 & 45.62 & 22.31 & 9.62 & 3.95 & 72.21 & 70.84 & 70.03 & 68.81\\
      \textbf{KL+R} & 10.34 & 78.44 & 46.95 & 23.80 & 11.06 & 4.58 & 73.36 & 72.39 & 71.79 & 70.23\\
      \hline
      RKL & 10.69 & 78.17 & 45.21 & 21.66 & 9.48 & 3.86 & 71.73 & 70.39 & 69.70 & 68.49\\
      \textbf{RKL+R} & 10.72 & 78.24 & 46.53 & 23.33 & 10.72 & 4.84 & 73.02 & 72.08 & 71.51 & 69.08\\
      \hline
      JSD & 10.35 & 78.54 & 45.85 & 22.59 & 10.16 & 4.16 & 72.29 & 70.92 & 70.15 & 68.89\\
      \textbf{JSD+R} & 10.76 & 78.25 & 46.72 & 23.81 & 11.14 & 4.67 & 73.40 & 72.63 & 72.10 & 70.22\\
      \hline
      TVD & 10.45 & 78.60 & 45.71 & 22.08 & 9.81 & 4.53 & 71.99 & 70.69 & 69.85 & 68.56\\
      \textbf{TVD+R} & 10.71 & 78.31 & 47.02 & 23.65 & 10.90 & 5.14 & 73.31 & 72.58 & 72.01 & 70.32\\
      \hline
      AKL & 10.34 & 78.35 & 45.69 & 22.42 & 10.05 & 3.99 & 72.25 & 70.91 & 70.18 & 68.97\\
      \textbf{AKL+R} & 10.44 & 78.35 & 46.81 & 23.59 & 10.89 & 4.63 & 73.21 & 72.28 & 71.70 & 70.16\\
      \hline
    \end{tabular}
    \caption{Multi-modal distribution learning situation on the pre-training task test set, which contains 5,000 slices.}
    \label{ood_pretrain}
  \end{center}
\end{table*}

\section{Validation Experiment of Different Ranking Objectives}
\label{svsp}

We have explained in Section \hyperref[sec4]{4} the reason for choosing SRCC instead of the Pearson correlation coefficient as the optimization objective for ranking loss. Because SRCC is more suitable for calculating ranking loss in scenarios involving discrete and non-linear language model output logits. 

In this section, we demonstrate the performance differences between applying these two sorting objectives on the GSM8K dataset (with the same experimental setup as in the main text) to show that SRCC indeed performs better in practical applications.

\begin{table*}[ht]
\small
\centering
\begin{tabular}{cccccccccc}
\hline
\textbf{Objective} & Rank-5 & Rank-15 & KL+R & RKL+R & JSD+R & TVD+R & FD+R & AKL+R & \textbf{Avg.}\\
\hline
Pearson & 246 & 229 & 230 & 147 & 233 & 202 & 250 & 223 & 220.00 \\
\textbf{SRCC} & 229 & 236 & 267 & 191 & 227 & 240 & 265 & 235 & \textbf{236.25} \\
\hline
\end{tabular}

\caption{
The number of correct instances in the test set of GSM8K when using different ranking consistency computation objectives as the optimization target.
}

\label{srcc_p}
\end{table*}

According to the results shown in Table \hyperref[srcc_p]{8}, the average performance of SRCC is better than that of the Pearson correlation coefficient, which is consistent with our theoretical estimation.

\section{Case Study}
\label{D}

In this section, we conduct a case study based on our experimental results on GSM8K. We select several cases and demonstrate the generated results before and after introducing ranking loss to some distillation objectives, as shown in Table \hyperref[tab6]{9}, Table \hyperref[tab7]{10} and Table \hyperref[tab8]{11}. 

Based on the results, we can see that after adding the ranking loss, the student model's choice of words and reasoning align more closely with the teacher model when answering questions. This also means that the peak predictions of the student model and the teacher model are more consistent.

Additionally, before introducing the ranking loss, TVD does not allow the student models to learn the answering pattern of GSM8K well, leading to automated evaluations failing to match the answers and resulting in a score of 0. After introducing the ranking loss, this deficiency is significantly improved, resulting in a good score.

\begin{table*}[ht]
\centering
\begin{tabular}{c|l}
\Xhline{3\arrayrulewidth}
\textbf{Question} & \makecell[l]{Greta and Celinda are baking cookies. Greta bakes 30 cookies and Celinda bakes \\twice as many. If the pair eat 10 of the cookies while they are cooling and put \\the rest in a box, how many cookies are there in the box?} \\
\Xhline{3\arrayrulewidth}
Ground truth & \makecell[l]{Celinda bakes 30 cookies * 2 = <<30*2=60>>60 cookies. So in total, Celinda \\and Greta bakes a total of 30 + 60 = <<30+60=90>>90 cookies. After eating \\some of the cookies, they have 90 baked cookies 10 eaten cookies = \\<<90-10=80>>80 cookies in the box.  \#\#\#\# 80} \\
\hline
Teacher & \makecell[l]{Celinda bakes 30 x 2 = <<30*2=60>>60 cookies. The total number of cookies is 30 \\+ 60 = <<30+60=90>>90 cookies. After they eat 10, there are 90 - 10 = <<90-10=80>>80 \\cookies in the box.  \#\#\#\# 80} \\
\hline
KL  & \makecell[l]{Celinda bakes 30 x 2 = <<30*2=60>>60 cookies. The total number of cookies baked is \\30 + 60 = <<30+60=90>>90 cookies. Greta and Celinda eat 10 + 10 = <<10+10=20>>20 \\cookies. There are 90 - 20 = <<90-20=70>>70 cookies in the box.  \#\#\#\# 70} \\
\hline
\textbf{KL+R} & \makecell[l]{Greta bakes 30 cookies and Celinda bakes twice as many, so Celinda bakes \\2*30 = <<30*2=60>>60 cookies. The total number of cookies baked is 30+60 = \\<<30+60=90>>90 cookies. The pair eats 10 of the cookies while they are cooling, so \\there are 90-10 = <<90-10=80>>80 cookies in the box.  \#\#\#\# 80} \\
\hline
RKL & \makecell[l]{Celinda bakes 30 * 2 = <<30*2=60>>60 cookies. The total number of cookies baked is \\30 + 60 = <<30+60=90>>90 cookies. The number of cookies that the pair eat is 30 + 10 \\= <<30+10=40>>40 cookies. There are 90 - 40 = <<90-40=50>>50 cookies left in the box.  \\\#\#\#\# 50}\\
\hline
\textbf{RKL+R} & \makecell[l]{Greta bakes 30 cookies and Celinda bakes 2*30 = <<30*2=60>>60 cookies. \\The total number of cookies is 30+60 = <<30+60=90>>90 cookies. The pair eat 10 \\of the cookies while they are cooling so there are 90-10 = <<90-10=80>>80 cookies in \\the box.  \#\#\#\# 80}\\
\hline
TVD & \makecell[l]{Celinda bakes 30 x 2 = <<30*2=60>>60 cookies. The total number of cookies baked is \\30 + 60 = <<30+60=90>>90 cookies. The total number of cookies in the box is 90 x 10 \\= <<90*10=900>>900 cookies.}\\
\hline
\textbf{TVD+R} & \makecell[l]{Greta bakes 30 cookies and Celinda bakes twice as many, so Celinda bakes \\2*30 = <<2*30=60>>60 cookies. The total number of cookies baked is 30+60 \\= <<30+60=90>>90 cookies. The pair eat 10 of the cookies while they are cooling, \\so there are 90-10 = <<90-10=80>>80 cookies in the box.  \#\#\#\# 80}\\
\Xhline{3\arrayrulewidth}
\end{tabular}
\caption{
Case 1. In this case, the thought processes of the existing distillation objectives are incorrect, leading to the generation of incorrect calculation formulas. The introduction of the ranking loss corrects the faulty calculation thinking, making the distillation objectives closer to the teacher model.
}
\label{tab6}
\end{table*}

\begin{table*}
\centering
\begin{tabular}{c|l}
\Xhline{3\arrayrulewidth}
\textbf{Question} & \makecell[l]{Rani is obsessed with sports cars. She wonders what the faster car ever made \\can go so she looks it up. What she finds out is that the fastest car was 1.25 \\times faster than the 2nd fastest car. The 2nd fastest car was 5 times faster than \\the average car. The average car can go 120 Miles per hour. How fast does \\the faster car go?} \\
\Xhline{3\arrayrulewidth}
Ground truth & \makecell[l]{The 2nd fastest car can go 600 MPH because 5 x 120 = <<5*120=600>>600  \\The faster car can go 750 MPH because 1.25 x 600 = <<1.25*600=750>>750    \#\#\#\# 750} \\
\hline
Teacher & \makecell[l]{The 2nd fastest car goes 120*5=<<120*5=600>>600 mph  \\The faster car goes 600*1.25=<<600*1.25=750>>750 mph    \#\#\#\# 750} \\
\hline
KL  & \makecell[l]{The 2nd fastest car is 1.25 times faster than the average car so it goes 1.25*120 = \\<<1.25*120=150>>150 Miles per hour  \\The average car can go 120 Miles per hour so the faster car goes 150-120 = \\<<150-120=30>>30 Miles per hour    \#\#\#\# 30} \\
\hline
\textbf{KL+R} & \makecell[l]{The 2nd fastest car is 5 times faster than the average car, so it goes 120 Miles \\per hour * 5 = <<120*5=600>>600 Miles per hour.  \\The fastest car is 1.25 times faster than the 2nd fastest car, so it goes 600 Miles \\per hour * 1.25 = <<600*1.25=750>>750 Miles per hour.    \#\#\#\# 750} \\
\hline
RKL & \makecell[l]{The 2nd fastest car was 5*120=<<5*120=600>>600 Miles per hour. \\The fastest car was 600*1.25=<<600*1.25=750>>750 Miles per hour.  The faster car was \\750-120=<<750-120=630>>630 Miles per hour.    \#\#\#\# 630}\\
\hline
\textbf{RKL+R} & \makecell[l]{The 2nd fastest car was 5 times faster than the average car, so it was 5*120 = \\<<5*120=600>>600 Miles per hour.  The fastest car was 1.25 times faster than the \\2nd fastest car, so it was 1.25*600 = <<1.25*600=750>>750 Miles per hour.    \#\#\#\# 750}\\
\hline
TVD & \makecell[l]{The 2nd fastest car was 5 times faster than the average car so it was 5*120 = \\<<5*120=600>>600 Miles per hour  The fastest car was 1.25 times faster than the \\2nd fastest car so it was 1.25*600 = <<1.25*600=750>>750 Miles per hour}\\
\hline
\textbf{TVD+R} & \makecell[l]{The 2nd fastest car is 5 times faster than the average car, so it can go 120 \\Miles per hour * 5 = <<120*5=600>>600 Miles per hour.  The fastest car is 1.25 \\times faster than the 2nd fastest car, so it can go 600 Miles per hour * 1.25 =\\ <<600*1.25=750>>750 Miles per hour.    \#\#\#\# 750}\\
\Xhline{3\arrayrulewidth}
\end{tabular}
\caption{
Case 2. In this case, both KL and RKL lead to incorrect understanding of the problem statement, which is successfully corrected after introducing the ranking loss. Although TVD arrives at the correct answer, it does not know how to output it in the GSM8K format. After introducing the ranking loss, it can output the result in the correct format.
}
\label{tab7}
\end{table*}

\begin{table*}
\centering
\begin{tabular}{c|l}
\Xhline{3\arrayrulewidth}
\textbf{Question} & \makecell[l]{An 8-year old child wants to buy a toy car which costs \$12. He already has \\\$4 savings. How many days will it take him to save the remaining amount \\of money if he promises to save \$2 daily from his allowance?} \\
\Xhline{3\arrayrulewidth}
Ground truth & \makecell[l]{The boy still needs \$12 - \$4 = \$<<12-4=8>>8. If he saves \$2 per day then it will \\take him \$8 / \$2/day = <<8/2=4>>4 days to save the remaining amount. \#\#\#\# 4} \\
\hline
Teacher & \makecell[l]{The child needs to save \$12 - \$4 = \$<<12-4=8>>8 more. He will need \$8 / \$2 = \\<<8/2=4>>4 more days to save up enough money.\#\#\#\# 4} \\
\hline
KL  & \makecell[l]{The child needs to save \$12 - \$4 = \$<<12-4=8>>8 more. He will save \$8/day * 7 days =\\ \$<<8*7=56>>56 in total. It will take him \$56 / \$2/day = <<56/2=28>>28 days to save the \\remaining amount of money.\#\#\#\# 28} \\
\hline
\textbf{KL+R} & \makecell[l]{The child needs to save \$12 - \$4 = \$<<12-4=8>>8 more. He will save \$8 / \$2 =\\ <<8/2=4>>4 days.\#\#\#\# 4} \\
\hline
RKL & \makecell[l]{The child will save \$12 - \$4 = \$<<12-4=8>>8 in 8 days. It will take him 8 days to \\save \$8.\#\#\#\# 8}\\
\hline
\textbf{RKL+R} & \makecell[l]{The child needs to save \$12 - \$4 = \$<<12-4=8>>8. He will have to save for \\8 / 2 = <<8/2=4>>4 days.\#\#\#\# 4}\\
\hline
TVD & \makecell[l]{The child will need to save \$12 - \$4 = \$<<12-4=8>>8. It will take him 8 / 2 = \\<<8/2=4>>4 days to save the remaining amount of money.}\\
\hline
\textbf{TVD+R} & \makecell[l]{The child needs to save \$12 - \$4 = \$<<12-4=8>>8. He will save \$8 / \$2 = \\<<8/2=4>>4 days.\#\#\#\# 4}\\
\Xhline{3\arrayrulewidth}
\end{tabular}
\caption{
Case 3. As same as Case 2, in this case, both KL and RKL lead to incorrect understanding of the problem statement, which is corrected after introducing the ranking loss. Moreover, TVD can output the result in the correct format after introducing the ranking loss.
}
\label{tab8}
\end{table*}

\end{document}